\documentclass{article}
\usepackage{spconf,amsmath,epsfig}

\usepackage{xspace}
\usepackage{algorithm}
\usepackage{graphicx} 
\usepackage{algorithmic}
\usepackage{bm}
\usepackage{mathrsfs}
\usepackage{enumitem}
\usepackage{amsmath}

\usepackage{amssymb}
\usepackage{booktabs}
\usepackage{pifont}
\usepackage{multirow}
\usepackage{cancel}
\usepackage{dsfont}
\usepackage{subfig}
\usepackage{marvosym}
\newcommand{\etal}{\textit{et al}.}

\newcommand{\ie}{\textit{i}.\textit{e}.}

\newcommand{\ourmethod}{\textsc{CMAH}\xspace}
\newcommand{\cmark}{\ding{51}}
\newcommand{\xmark}{\ding{55}}

\let\OLDthebibliography\thebibliography
\renewcommand\thebibliography[1]{
  \OLDthebibliography{#1}
  \setlength{\parskip}{0pt}
  \setlength{\itemsep}{0pt plus 0.3ex}
}

\begin{document}\sloppy

\def\x{{\mathbf x}}
\def\L{{\cal L}}

\title{Contrastive masked auto-encoders based self-supervised hashing for 2D image and 3D point cloud cross-modal retrieval}


\name{Rukai Wei$^{1 \dag}$, Heng Cui$^{3\dag}$\thanks{$\dag$ Rukai Wei and Heng Cui contribute equally to this paper}, Yu Liu$^2$$*$\thanks{$*$ Yu Liu is the corresponding author at liu\_yu@hust.edu.cn}, Yufeng Hou$^{2}$, Yanzhao Xie$^4$, Ke Zhou$^1$}
\address{
$^1$Wuhan National Laboratory for Optoelectronics, Huazhong University of Science and Technology\\$^2$School of Computer
Science and Technology, Huazhong University of Science and Technology\\$^3$School of Software Engineering, Huazhong University of Science and Technology\\$^4$School of Computer Science and Cyber Engineering, Guangzhou University}

\maketitle
%
\begin{abstract}
Implementing cross-modal hashing between 2D images and 3D point-cloud data is a growing concern in real-world retrieval systems. Simply applying existing cross-modal approaches to this new task fails to adequately capture latent multi-modal semantics and effectively bridge the modality gap between 2D and 3D. To address these issues without relying on hand-crafted labels, we propose \textbf{\underline{c}}ontrastive \textbf{\underline{m}}asked \textbf{\underline{a}}utoencoders based self-supervised \textbf{\underline{h}}ashing (\ourmethod) for retrieval between images and point-cloud data. We start by contrasting 2D-3D pairs and explicitly constraining them into a joint Hamming space. This contrastive learning process ensures robust discriminability for the generated hash codes and effectively reduces the modality gap. Moreover, we utilize multi-modal auto-encoders to enhance the model's understanding of multi-modal semantics. By completing the masked image/point-cloud data modeling task, the model is encouraged to capture more localized clues. In addition, the proposed multi-modal fusion block facilitates fine-grained interactions among different modalities. Extensive experiments on three public datasets demonstrate that the proposed \ourmethod significantly outperforms all baseline methods.

\end{abstract}
\section{Introduction}
\label{sec:intro}

3D point-cloud data captures the spatial layout and geometric properties of objects or scenes within a three-dimensional space. In real-world applications like autonomous driving, augmented reality, and robotics, retrieving 3D point-cloud data from 2D images (or vice versa) quickly is crucial. With the advantage of fast retrieval speed, cross-modal hashing~\cite{DGCPN2021AAAI,UCCH2023TPAMI,DUMCH2023ICME} can convert high-dimensional multi-modal data into binary hash codes that are used to retrieve data by their Hamming distances. Previous research has shown promising results in image-text and image-video scenarios. However, these findings may not apply to the emerging 3D-2D scenario.

\noindent\textbf{Motivation:} Since 3D point-cloud data~\cite{STAEM2023ICME} differs significantly from 2D images in their formation and description, imposing existing self-supervised cross-modal hashing methods to the retrieval task in this scenario inevitably poses the challenge of effectively bridging the modality gap between images and point-cloud data. First, the irregular and unordered data structure of point-cloud data makes it difficult to capture meaningful semantics effectively. As a result, the generated hash codes struggle to accurately preserve data relationships from the original space into the Hamming space. Second, the feature variation and semantic gap between 2D pixels and 3D coordinates prohibit the learning of accurate correspondence between two modalities. This leads to the hash codes of different modalities being widely dispersed in the Hamming space.  To address these challenges, we are motivated to find solutions from the perspectives of local feature capture and global relationship preservation, respectively.

\noindent\textbf{Our Approach:} To this end, we incorporate multi-modal contrastive learning and masked image/point-cloud modeling. The former is good at preserving global relationships, while the latter specializes in capturing local features.

Specifically, we utilize a contrastive learning model to align modalities between 2D images and 3D point clouds (see $\S~\ref{sec:contrastive_learning}$). Instead of directly contrasting full data, we contrast both full-full and masked-full pairs of the image and point-cloud data. This approach generates modality-invariant hash codes that can reduce noise from local redundancies. The contrastive learning task helps the model learn a shared Hamming space where hash codes with similar semantics from different modalities can be closely mapped together, effectively reducing the modality gap.

In addition, we introduce a masked auto-encoder model to capture detailed information in images and point-cloud data (see $\S~\ref{sec:masked_autoencoder}$). The model first encodes unmasked patches into a latent space and then decodes all patches to reconstruct masked patches. This process encourages the model to gain comprehensive knowledge about the local 2D-3D context. Furthermore, we observed that the information captured from a modality can serve as a valuable complementary prompt to capture accurate information from the other modality that is masked. Therefore, we incorporate a multi-modal fusion block between the encoders and decoders. This allows for fine-grained interaction between the image and point-cloud data, facilitating the capture of more localized associations.

As a result, we propose \textbf{\underline{c}}ontrastive  \textbf{\underline{m}}asked \textbf{\underline{a}}uto-encoders based self-supervised \textbf{\underline{h}}ashing (\ourmethod) for image and point-cloud cross-modal retrieval. \ourmethod is capable of extracting meaningful semantics as well as reducing the modality gap, thereby generating discriminative hash codes. We conducted extensive experiments on three benchmark datasets, and the results demonstrate that \ourmethod can achieve remarkable cross-modal retrieval performance. 

\noindent\textbf{Contributions:}

\begin{itemize}
    \item To the best of our knowledge, this is the first job that attempts to address the issue of self-supervised cross-modal hashing for images and point-cloud data. Our proposed contrastive masked auto-encoders based hashing (\ourmethod) achieves state-of-the-art cross-modal retrieval performance in this field.
    \item We utilize multi-modal contrastive learning to align image and point-cloud data, resulting in the explicit reduction of the modality gap in Hamming space and generating distinguishable hash codes.
    \item We propose utilizing masked auto-encoders and incorporating a multi-modal fusion block to comprehensively understand multi-modal semantics, resulting in accuracy in feature capture from fine-grained interactions between images and point-cloud data by localized associations.
\end{itemize}

\section{Related Work}

\noindent\textbf{Self-supervised cross-modal hashing.}
Self-supervised cross-modal hashing~\cite{DJSRH2019ICCV,JDSH2020SIGIR,DGCPN2021AAAI,CMFH2014CVPR,UGACH2018AAAI} has gained increasing attention and shown promising performance in retrieval systems for cross-modal retrieval among image, text, and video. Existing methods focus on the generation of supervisory signals and thereby achieve high-quality hash codes. For example, DJSRH~\cite{DJSRH2019ICCV} and JDSH~\cite{JDSH2020SIGIR} investigated similarity fusion to construct multi-modal pairwise similarities as pseudo labels. DGCPN~\cite{DGCPN2021AAAI} and ASSPH~\cite{ASSPH2022MM} introduce a deep graph-neighbor coherence-preserving network and an adaptive learning scheme, respectively, to enhance the learning of more precise structural correlations. However, none of them are specifically designed for the cross-modal retrieval of images and point-cloud data. Adapting these methods to this new task is costly and yields sub-optimal results.

\noindent\textbf{Self-supervised representation learning.} Self-supervised representation learning serves as a foundational aspect for numerous downstream tasks. In the context of image representation learning, methods like SimCLR~\cite{SimCLR2023ICML}, MoCo~\cite{MoCo2020CVPR}, and PointContrast~\cite{PointContrast2020ECCV} \etal~have been proposed to learn transferable representations by leveraging the contrastive learning framework to attract positive pairs and repel negative ones. In addition, MAE~\cite{MAE2022CVPR} and  Point-MAE~\cite{PointMAE2022ECCV} employ auto-encoders to train a robust vision learner through the masked image modeling task. In the field of multi-modal representation learning, the CLIP~\cite{CLIP2021ICML} model, pre-trained on large-scale image-text datasets, has exhibited remarkable capabilities for zero-shot recognition. Furthermore, models like ALBEF~\cite{ALBEF2021NIPS}, CoCa~\cite{CoCa2022TMLR} \etal~incorporate multiple learning tasks and have achieved promising results, which have inspired the design of our proposed \ourmethod.

\begin{figure*}[t]
    \centering
    \includegraphics[width=0.85\textwidth]{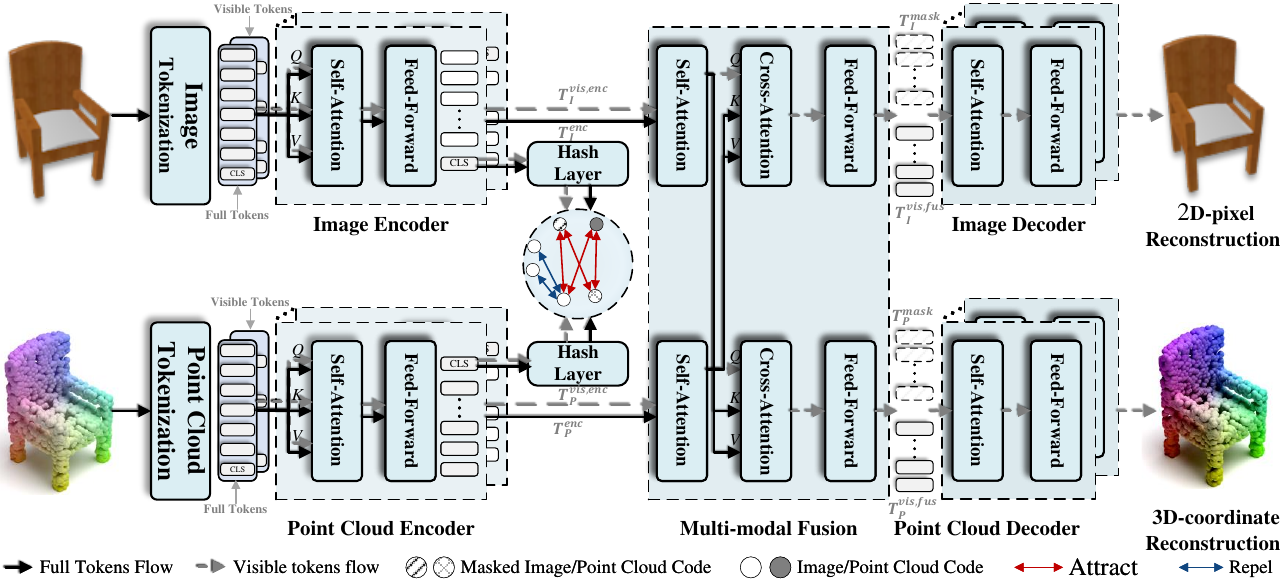}
    \caption{The overall framework of \ourmethod. On the one hand, both masked visible tokens and full tokens are encoded and projected as hash codes. Contrastive learning is conducted using both full-full and mask-full pairs. On the other hand, the encoded tokens are inputted into a multi-modal fusion block to facilitate fine-grained multi-modal interaction. Subsequently, they are directed to their respective decoders for 2D-pixel and 3D-coordinate reconstructions, respectively.}
    \label{fig:framework}
    \vspace{-1.5em}
\end{figure*}

\section{Methodology}
\subsection{Problem definition $\&$ Overview}

Given a dataset $\mathcal{D}=\left\{\left(P_{i},I_{i}\right)\right\}_{i=1}^{|\mathcal{D}|}$ with $P_{i}\in \mathbb{R}^{N\times3}$ and $I_{i}\in \mathbb{R}^{H\times W \times 3}$, where $N$ is the number of points in point-cloud data and $H \times W$ denotes the image size. Note that $I_{i}$ is obtained by capturing $P_{i}$ from a random camera viewpoint. The proposed \ourmethod aims to learn non-linear hash functions that map $P_{i}$ and $I_{i}$ to compact hash codes $b_{i}^{P} \in \left\{-1,1\right\}^{K}$ and $b_{i}^{I} \in  \left\{-1,1\right\}^{K}$, respectively, where $K$ is the code length. Note that both point-cloud data and images are unannotated. Thus, \ourmethod learns hash functions without the aid of hand-crafted labels. The framework of \ourmethod is displayed in Figure~\ref{fig:framework}.

\noindent\textbf{Design of \ourmethod modules.} Since the Transformer~\cite{transformer2017NIPS} model has gained success in many vision tasks, we adopt transformer-based models to encode both point-cloud data and images. For the former, we adopt the approach described in Point-MAE~\cite{PointMAE2022ECCV}, where it employs Furthest Point Sampling (FPS) and $k$-nearest neighbors ($k$-NN) for downsampling. We also utilize a mini-PointNet for point-cloud tokenization and yield $M_{P}$ tokens, which are denoted as $T_{P} \in \mathbb{R}^{M_{P}\times 384}$, where $M_{P}=64$. In this way, each token dominates a local spatial region and interacts with others in the subsequent attention module, enabling the integration of wide-range features. For the latter, we follow ViT~\cite{ViT2021ICLR} to divide an image into patches and embed them into $M_{I}$ image tokens, which are denoted as $T_{I} \in \mathbb{R}^{M_{I}\times 768}$, where $M_{I}=196$. As shown in Figure~\ref{fig:framework}, those encoders consist of multiple transformer blocks, each of which consists of a multi-head self-attention module and a feed-forward network. Furthermore, we adopt the cross-attention mechanism to fuse 2D–3D latent features in the multi-modal fusion block and append the feed-forward network for feature projection. To complete the mask modeling task, we also deploy a transformer-based decoder for reconstruction, which is similar to the encoder but contains fewer transformer blocks. 

\noindent\textbf{Process of \ourmethod optimization.} During the training phase, the full images and point-cloud data are initially tokenized into $T_{P}$ and $T_{I}$, respectively. We randomly mask a significant portion of tokens, leaving only the visible tokens $T_{I}^{vis} \in \mathbb{R}^{M_{I}^{vis}\times 768}$ and $T_{P}^{vis} \in \mathbb{R}^{M_{P}^{vis}\times 384}$. The forward propagation of both the image and point-cloud encoders runs twice: once for full tokens in $T_{P}$ and $T_{I}$, and another time for the visible tokens $T_{I}^{vis}$ and $T_{P}^{vis}$. Then, the output embeddings derived from [$CLS$] tokens will be projected into continuous hash codes $h^{P}$ and $h^{I}$ through the hash layer, respectively. All hash codes converted from two modalities will be used for contrastive learning to compute contrastive loss (see $\S~\ref{sec:contrastive_learning}$ for details). For the remaining output visible token embeddings $T_{I}^{vis, enc}$ and $T_{P}^{vis, enc}$ (encoded by $T_{I}^{vis}$ and $T_{P}^{vis}$), they will be fused with $T_{P}^{enc}$ and $T_{I}^{enc}$ (encoded by $T_{I}$ and $T_{P}$) in the multi-modal fusion block, respectively. Furthermore, we concatenate the fused visible embeddings with a set of masked tokens that are shared and learnable and input them into lightweight transformer-based decoders. Subsequently, the decoded masked tokens of the image and point-cloud data are employed to compute reconstruction losses that are specific to each modality. The hash function is optimized via the simultaneous application of both contrastive and reconstructive objectives. During the test phase, we will deactivate the multi-modal fusion block and the decoders. Instead, we will employ the corresponding encoders and hash layers, along with the $sign$ function for binarization, to convert full images and point-cloud data into binary codes.

\subsection{Multi-modal contrastive learning}
\label{sec:contrastive_learning}
The contrastive learning between 2D-3D pairs can effectively reduce the modality gap and assist in capturing the semantic structure of images and point-cloud data. In particular, instance-discrimination-based contrastive learning~\cite{SimCLR2023ICML} aims to maximize the similarity between positive pairs while minimizing the similarity between negative pairs. It achieves this goal by minimizing the contrastive loss function
\begin{equation}
    \begin{aligned}
        \ell(x,x^{+})\!=\!\!-\!\log \frac{e^{sim(x,x^{+})/\tau}}{e^{sim(x,x^{+})/\tau}+\sum_{x^{-}\in \mathcal{N}} e^{sim(x,x^{-})/\tau}},
    \end{aligned}
\end{equation}
where $x^{+}$ and $x^{-}$ denote the positive and negative pairs, respectively, associating with the anchor sample $x$. $\mathcal{N}$ denotes the set of negative samples for the anchor sample $x$, and $\tau$ is the temperature parameter.

If a pair $(P_{i},I_{i})$ experiences the masked, encoded, and projected into hash codes operations, we will obtain four continuous hash codes: $h_{i}^{P}$, $h_{i}^{I}$, $h_{i}^{P, vis}$, and $h_{i}^{I, vis}$. These hash codes correspond to the full point-cloud data, full image, masked point-cloud data, and masked image, respectively. To contrast between full images and full point-cloud data, we formulate the contrastive loss below.
\begin{equation}
\begin{aligned}
L_{f-f} =\frac{1}{2 \cdot B}\sum_{i=1}^{B} \ell(h_{i}^{P},h_{i}^{I})+\ell(h_{i}^{I},h_{i}^{P}),
\end{aligned}
\end{equation}
where $B$ represents the batch size. Note that we define point-cloud data and its rendered image as constituting a positive pair. Meanwhile, all samples within a mini-batch, excluding the positive sample, are considered negative samples for the anchor sample. For instance, when contrasting $\left\{h_{i}^{P}\right\}_{i=1}^{B}$ and $\left\{h_{i}^{I}\right\}_{i=1}^{B}$, given the anchor $h_{i}^{P}$, the negative sample set is defined as $\left\{h_{j}^{I}\right\}_{j=1, j\neq i}^{B} \cup \left\{h_{j}^{P}\right\}_{j=1, j\neq i}^{B}$.

Furthermore, recent studies~\cite{ConMH2023AAAI,CMAE2022arxiv} have demonstrated the effectiveness of masking as an augmentation technique in contrastive learning. By eliminating local redundancy, masking can enhance the model's robustness and generation ability. Therefore, we supplement the contrast between pairs consisting of masked images and full point-cloud data and pairs consisting of full images and masked point-cloud data. The contrastive loss for the former pairs is defined as
\begin{equation}
\begin{aligned}
L_{f-m}=\frac{1}{2\cdot B}\sum_{i=1}^{B} \ell(h_{i}^{P},h_{i}^{I, vis})+\ell(h_{i}^{I,vis},h_{i}^{P}),
\end{aligned}
\end{equation}
while the contrastive loss for the latter pairs is defined as
\begin{equation}
\begin{aligned}
L_{m-f}=\frac{1}{2\cdot B}\sum_{i=1}^{B} \ell(h_{i}^{P,vis},h_{i}^{I})+\ell(h_{i}^{I},h_{i}^{P,vis}).
\end{aligned}
\end{equation}

Note that \ourmethod does not perform contrastive learning between masked images and masked point-cloud data because masking both the image and point-cloud data could result in the masked pairs no longer representing the same semantic. This contrast could potentially introduce false positive pairs. It will diminish the discrimination ability of the hashing model and adversely affect the final retrieval performance. Finally, the overall contrastive loss is formulated as
\begin{equation}
\begin{aligned}
L_{c}=L_{f-f}+L_{f-m}+L_{m-f}.
\end{aligned}
\end{equation}

\subsection{Masked auto-encoders with multi-modal fusion}
\label{sec:masked_autoencoder}
While the above learning task can aid the model in capturing the semantic structure between samples across modalities and explicitly reducing the modality gap, it primarily focuses on modeling the relationships among different samples. Local 2D–3D cues are also worth leveraging for semantic hash learning. In this section, we will describe the design of masked auto-encoders with multi-modal fusion for local perception.

\noindent\textbf{Multi-modal fusion block.}
To combine contextualized representations from both images and point-cloud data, we incorporate a co-attention mechanism within the multi-modal fusion module. This module consists of a self-attention layer, a cross-attention layer, and a feed-forward layer. It facilitates the integration of information across modalities. Implementing the encoding process, we can obtain the encoded tokens $T_{I}^{vis,enc}$, $T_{P}^{vis,enc}$, $T_{I}^{enc}$, and $T_{P}^{enc}$. Subsequently, we fuse $T_{P}^{enc}$ with $T_{I}^{vis,enc}$ and $T_{I}^{enc}$ with $T_{P}^{vis,enc}$ shown in Figure~\ref{fig:framework}. To elaborate on this process, in the point-cloud branch, we employ the entire set consisting of images and encoded tokens as prompts. $T_{I}^{enc}$ initially undergoes the self-attention layer and subsequently serves as both the `Key' and `Value' for the cross-attention mechanism. The output of $T_{P}^{vis,enc}$ from the self-attention layer is employed as the `Query'. $Q$, $K$, and $V$ will be sent to the cross-attention layer, where we can achieve fine-grained multi-modal interaction. Finally, the features yielded by the attention layer pass through the Feed-Forward Network (FFN) layer, and we obtain the fused feature $T_{P}^{vis, fus} \in \mathbb{R}^{M_{P}^{vis}\times 768}$. Similarly, in the image branch, we utilize the projection of $T_{P}^{enc}$ as both the `Key' and `Value' to obtain $T_{I}^{vis, fus} \in \mathbb{R}^{M_{I}^{vis}\times 768}$.


\noindent\textbf{Decoding and reconstruction.}
In the reconstruction phase, we concatenate the fused tokens from each modality with the corresponding set of shared learnable masked tokens, \ie, $T_{I}^{mask}\in \mathbb{R}^{M_{I}^{mask}\times 768}$ or $T_{P}^{mask}\in \mathbb{R}^{M_{P}^{mask}\times 384}$. $M^{mask}$ denotes the number of masked tokens, and $M^{mask}+M^{vis}=M$. In the transformer decoder, the masked tokens learn to capture informative spatial cues from visible ones and reconstruct the masked 3D coordinates and 2D patches. 

For point-cloud data reconstruction, we follow the previous study~\cite{PointMAE2022ECCV} to utilize the decoded masked tokens $T_{P}^{dec, mask}$ to reconstruct 3D coordinates of the masked tokens along with their $k$ neighboring points. We employ a reconstruction head comprising a single linear projection layer, \ie, $\mathcal{P}_{3D}$, to predict the ground-truth 3D coordinates of the masked points, \ie, $P^{mask} \in \mathbb{R}^{M_{P}^{mask}\times k \times 3}$. Then, we follow the study~\cite{PointMAE2022ECCV} to compute the loss by Chamfer Distance and formulate it as
\begin{equation}
\small
    \begin{aligned}
        L_{3D}\!\!=\!\!\frac{1}{B}\!\!\sum_{i=1}^{B}\! \frac{1}{M_{P,i}^{mask}k}Chamfer\!\!\left(\!\mathcal{P}_{3D}\!\!\left(\!T_{P,i}^{dec,mask}\!\right)\!,P_{i}^{mask}\!\right).
    \end{aligned}
\end{equation}

For image reconstruction, we follow the study~\cite{MAE2022CVPR} to reshape the decoder's output to form a reconstructed image and reconstruct the input by predicting the pixels for each masked patch. We adopt MSE as the 2D reconstruction loss. The loss is described as
\begin{equation}
\small
    \begin{aligned}
        L_{2D}\!\!=\!\!\frac{1}{B}\!\sum_{i=1}^{B}\!\frac{1}{M_{I,i}^{mask}}\! MSE \left(\!\mathcal{P}_{2D}\!\left(T_{I,i}^{dec,mask}\right),I_{i}^{mask}\right),
    \end{aligned}
\end{equation}
where $I_{i}^{mask}$ denotes the ground truth pixel values for the masked patches and $\mathcal{P}_{2D}$ is the projector. 

The overall reconstruction loss can be formulated as
\begin{equation}
    \begin{aligned}
        L_{r}=L_{3D}+L_{2D}.
    \end{aligned}
\end{equation}

Finally, the hash model is collaboratively optimized by minimizing reconstruction loss and contrastive loss. As a result, the overall learning objective can be defined as
\begin{equation}
    \begin{aligned}
        L_{overall}=L_{c}+L_{r}.
    \end{aligned}
\end{equation}

\section{Experiments}

\subsection{Experimental settings}
In the interest of saving space, we only give a brief description of the experimental settings. For more details, please refer to \textbf{supplementary materials}.

\noindent\textbf{Datasets.} Following the settings of previous studies~\cite{DGCPN2021AAAI, DJSRH2019ICCV, JDSH2020SIGIR}, we employ three publicly available datasets to validate the effectiveness of \ourmethod, including ShapeNetRender, ModelNet, and ShapeNet-55. For the first dataset, we select 5,000 pairs as the query set and 38,783 pairs as the retrieval set. From the retrieval set, we randomly select 10,000 pairs for training. We partition the second dataset into a query set comprising 2,000 pairs and a retrieval set containing 10,311 pairs. In the retrieval set, we utilize 5,000 pairs for training. In the case of the last dataset, we designate 5,000 pairs for the query set, while the remaining 47,460 pairs form the retrieval set. Specifically, we select 10,000 pairs from the retrieval set for training purposes.

\noindent\textbf{Baselines and evaluation metrics.}
We adopt the mean Average Precision (mAP) and precision@top-K curves to display the performance variation. Besides, we adapt several state-of-the-art image-text methods to achieve image-3D point cloud retrieval, including  DJSRH~\cite{DJSRH2019ICCV}, DGCPN~\cite{DGCPN2021AAAI}, ASSPH~\cite{ASSPH2022MM}. We compare their performance on both $I\rightarrow P$ and $P \rightarrow I$ retrieval tasks in terms of mAP@2000.

\noindent\textbf{Implementation details.} We configure the transformer encoders for both images and point-cloud data with 12 layers, while the number of layers of decoders is 8 and 4 for the image branch and the point-cloud branch, respectively. The image encoder-decoder is pre-trained on ImageNet, while the point-cloud encoder-decoder is pre-trained on ShapeNet. The masking ratio of an image is $75\%$, while the counterpart of point-cloud data is $60\%$. The hash layer comprises two fully-connected layers, with ReLU as the activation function and $tanh$ for continuous relaxation. For training, we set the batch size $B = 32$ and an initial learning rate of 0.0001. It will decay by $90\%$ every 20 epochs until reaching a minimum learning rate of 0.00001. The AdamW optimizer is utilized to optimize the objective with a momentum of $0.9$. Our method is implemented in PyTorch and is trained on a single NVIDIA RTX 3090 GPU.

\begin{table}[t]
    \large
    \centering
    \caption{The mAP@2000 results in $I\rightarrow P$ and $P \rightarrow I$ retrieval tasks on three public datasets.}
    \resizebox{\linewidth}{!}{
     \begin{tabular}{ccccccccccc}
    \toprule
    \multirow{2}{*}{Task}&
    \multirow{2}{*}{Method}&
    \multicolumn{3}{c}{ShapeNetRender}&
    \multicolumn{3}{c}{ModelNet}&
    \multicolumn{3}{c}{ShapeNet-55}\\
    \cmidrule(lr){3-5} \cmidrule(lr){6-8} \cmidrule(lr){9-11} 
    & &16-bit&32-bit&64-bit&16-bit&32-bit&64-bit&16-bit&32-bit&64-bit \\
    \midrule 
    \multirow{4}{*}{I$\rightarrow$P}
    &DJSRH\cite{DJSRH2019ICCV}&0.624&0.642&0.663&0.314&0.349&0.393&0.466&0.491&0.513\\
    &DGCPN\cite{DGCPN2021AAAI}&0.681&0.695&0.717&0.356&0.387&0.424&0.513&0.532&0.560\\
    &ASSPH\cite{ASSPH2022MM}&0.716 &0.740 &0.749&0.373&0.411&0.446&0.541&0.564&0.593\\
    
    &\textbf{\ourmethod}&\bf 0.760& \bf 0.775& \bf 0.791&\bf 0.409&\bf  0.466& \bf 0.501&\bf 0.577&\bf 0.619 &\bf 0.637 \\
    \midrule
    \midrule
    \multirow{4}{*}{P$\rightarrow$I}
    &DJSRH\cite{DJSRH2019ICCV}&0.610&0.647&0.661&0.357&0.384&0.429&0.474&0.506&0.531\\
    &DGCPN\cite{DGCPN2021AAAI}&0.674&0.697&0.711&0.386&0.422&0.465&0.522&0.549&0.576\\
     &ASSPH\cite{ASSPH2022MM}&0.709 &0.736 &0.744 &0.409 &0.448&0.487&0.557&0.581&0.614\\
    &\textbf{\ourmethod}&\bf 0.758&\bf 0.772&\bf 0.790&\bf 0.452&\bf 0.516&\bf 0.545& \bf 0.585 & \bf 0.626 & \bf 0.647 \\
    \bottomrule
    \end{tabular}
    }
    \label{tab:map}
    \vspace{-1em}
\end{table}

\subsection{Comparisons with SOTAs}

\noindent\textbf{mAP comparisons.} We first compare \ourmethod to baseline methods in terms of mAP@2000. The results are reported in Table~\ref{tab:map}. We can observe that \ourmethod outperforms other baselines on all three benchmark datasets at all code lengths. Specifically, \ourmethod achieves improvements of $5.6\%$, $12.3\%$, and $7.4\%$ compared to ASSPH on ShapeNetRender, ModelNet, and ShapeNet-55, respectively, on the $I\rightarrow P$ task at 64 bits. Similarly, on the $P\rightarrow I$ task, the improvements are $6.2\%$, $11.9\%$, and $5.4\%$, respectively. Furthermore, \ourmethod outperforms ASSPH significantly, even with low-bit hash codes. For instance, the performance improvements are up to $6.1\%$, $9.6\%$, and $6.6\%$ on the three datasets, respectively, at 16 bits on the $I\rightarrow P$ task. These results demonstrate our superiority in real-time scenarios that require high performance and low storage demands.

\begin{figure}[t]
    \centering
    
    \subfloat[$I\rightarrow P$ on ShapeNetRender]{\includegraphics[width=0.48\linewidth]{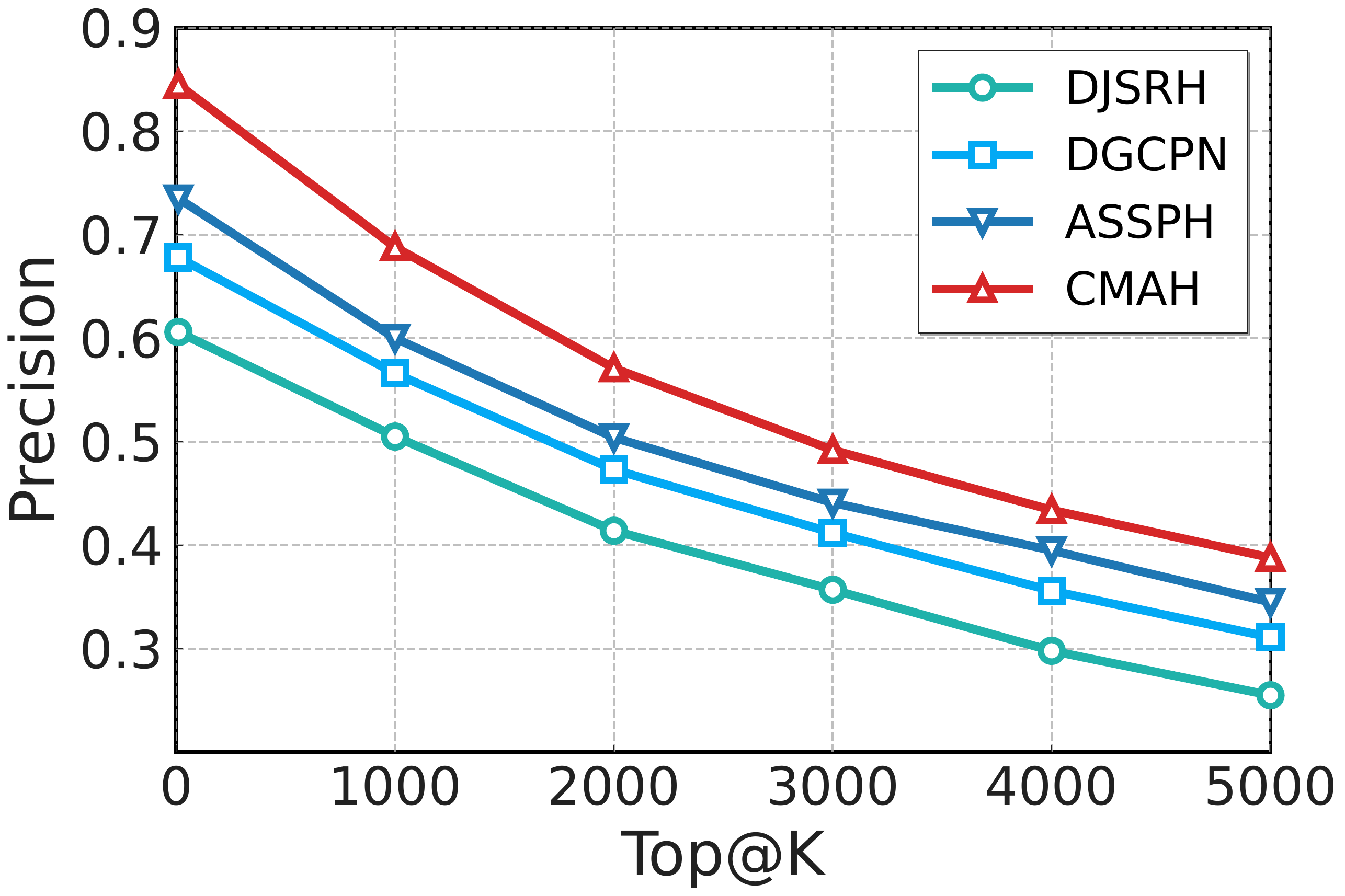}}
    \hfill
    \subfloat[$P\rightarrow I$ on ShapeNetRender]{\includegraphics[width=0.48\linewidth]{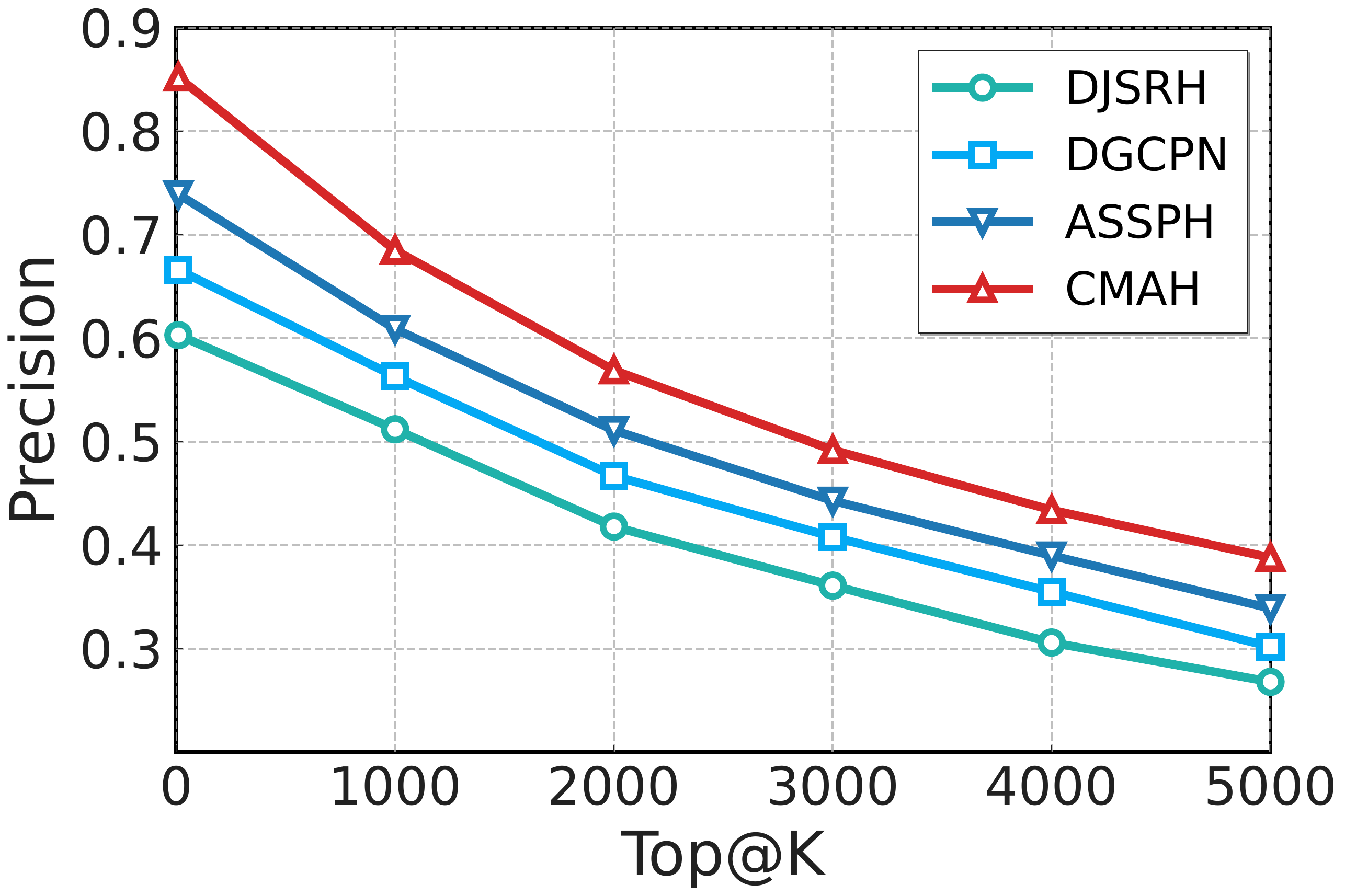}}\\
    \subfloat[$I\rightarrow P$ on ModelNet]{\includegraphics[width=0.48\linewidth]{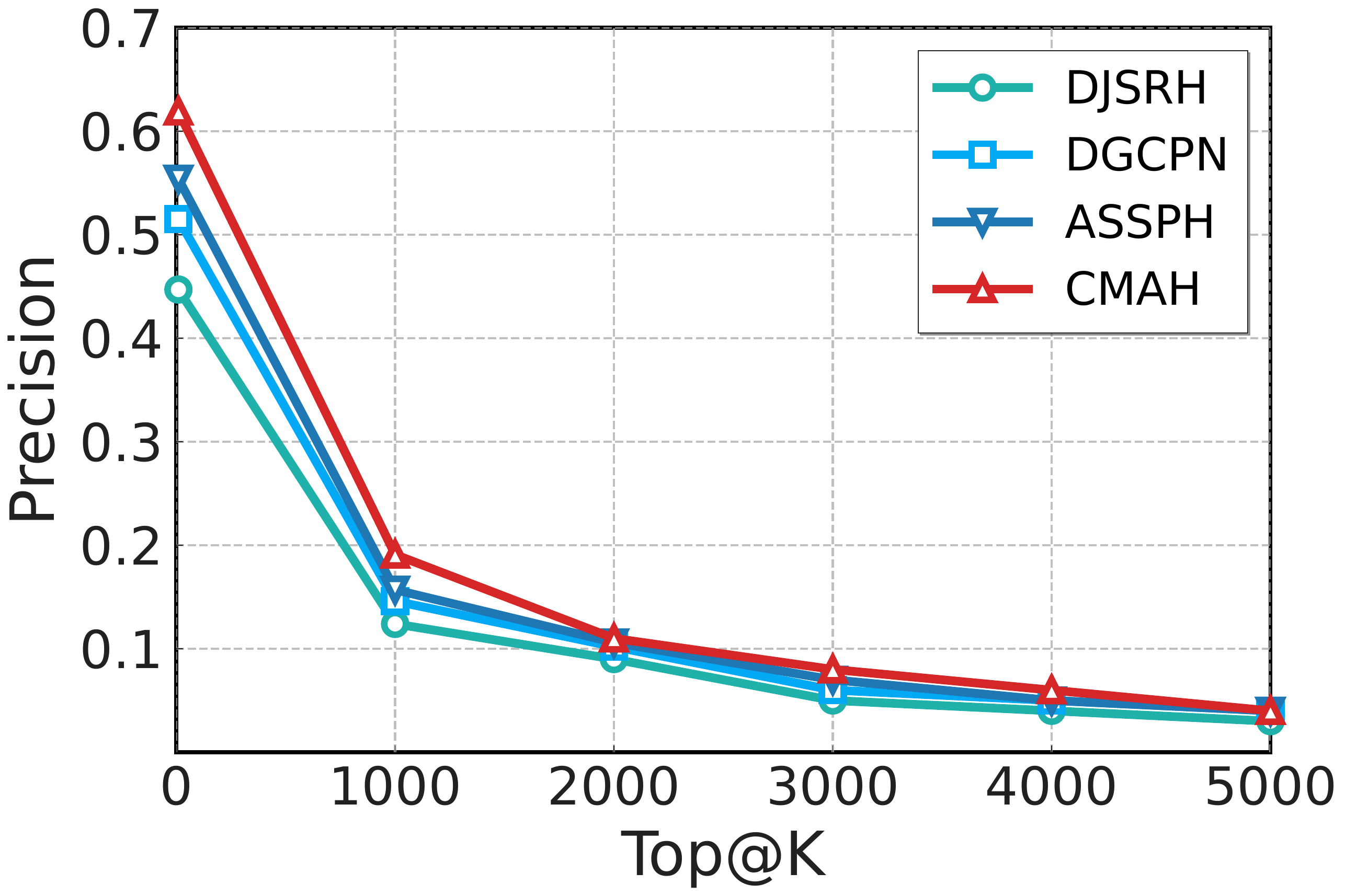}}
    \hfill
    \subfloat[$P\rightarrow I$ on ModelNet]{\includegraphics[width=0.48\linewidth]{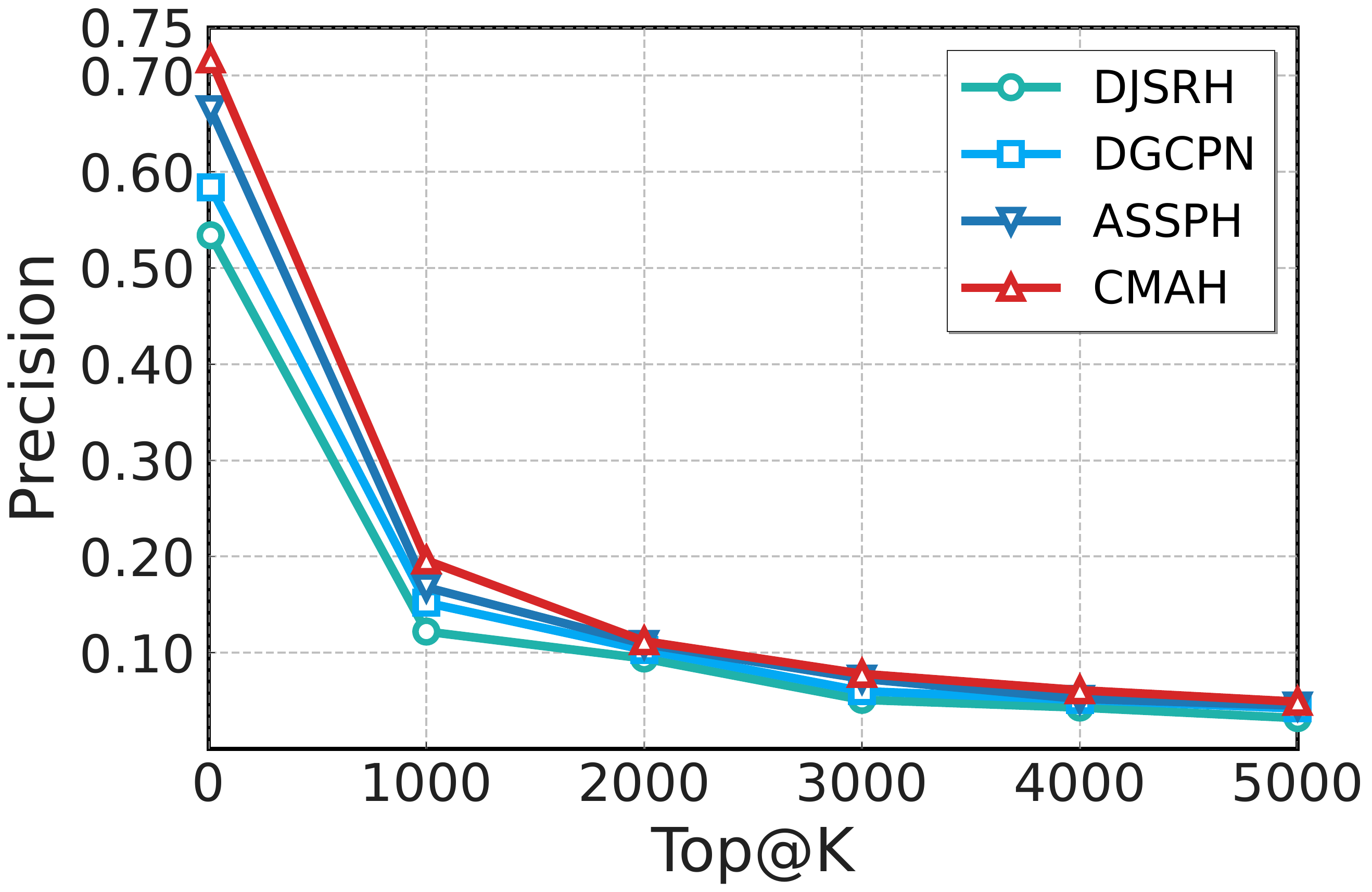}}
    \caption{The Precision@Top-K curves on different datasets at 64 bits.}
    \label{fig:precision}
    \vspace{-1.5em}
\end{figure}

\noindent\textbf{Precision@top-K comparisons.} We also display precision@top-K curves in Fig.~\ref{fig:precision}. The results demonstrate that, despite the decrease in precision exhibited by all methods as the search radius increases, \ourmethod consistently outperforms state-of-the-art baselines by significant margins across various datasets. This indicates that \ourmethod excels at retrieving more semantically similar and relevant results compared to others, irrespective of changes in the search radius.

These results demonstrate that our proposed hashing method effectively improves the quality of hash codes, resulting in optimal retrieval effects. We attribute it to \ourmethod successfully reducing the modality gap and preserving semantic information.

\subsection{Ablation study}
In this section, our objective is to examine the impact of each model component on the overall performance of cross-modal retrieval. The results are reported in Table~\ref{tab:ablation}.

\noindent\textbf{Effect of contrastive learning (CL).} We can observe from the first and seventh rows that our method experiences a huge drop in mAP performance when learned without the contrastive learning component. It is because the hash layer is only optimized by the contrastive learning objective. 
In addition, although both full-pair contrastive learning (second and eighth rows) and masked-pair contrastive learning (third and ninth rows) can achieve passable results, their collaboration can offer significant improvement. Specifically, the performance improvements on the ShapeNetRender and ModelNet datasets are as high as $2.39\%$ and $1.84\%$ on the $I \rightarrow P$ task, and $2.06\%$ and $2.51\%$ on the $P \rightarrow I$ task, respectively. These results indicate that CL is crucial for reducing the modality gap and capturing relationships between both inter-modality and intra-modality on cross-modal retrieval tasks for images and point-cloud data.

\noindent\textbf{Effect of mask reconstruction (MR).} We examine the impact of mask reconstruction in the fourth and tenth rows. Note that when we deactivate the MR module, the MF module is automatically disabled simultaneously. Comparing the learning results with MR in the sixth and twelfth rows, it is evident that there is an average mAP drop of $4.04\%$ and $4.82\%$ on ShapeNetRender and ModelNet, respectively, on the $I\rightarrow P$ task. Similarly, on the $P\rightarrow I$ task, the mAP drops by $4.09\%$ and $5.13\%$, respectively. Note that although this module does not directly affect the hash code, it is critical to the capability of perception of the encoder and is conducive to extracting informative semantic features. As a result, mask reconstruction is an important auxiliary task that empowers the model to focus more on local 2D-3D details for superior semantic understanding.

\noindent\textbf{Effect of multi-modal fusion (MF).} We investigate the impact of the proposed multi-modal fusion block in the third and sixth rows. We observe a slight performance decline on both the ShapeNetRender and ModelNet datasets. Based on these observations, we conclude that the multi-modal fusion block enhances the model's ability to capture more accurate multi-modal correspondences and achieve improved semantic understanding. This is achieved by enabling fine-grained interaction between modalities, using tokens from one modality as prompt information for reconstructing the other.

\begin{table}[t]
    \centering
    \caption{Ablation studies on ShapeNetRender and ModelNet to investigate the effectiveness of each model component. \textbf{CL-F}, \textbf{CL-M}, \textbf{MR}, and \textbf{MF} represent full-pair contrastive learning, masked-pair contrastive learning, mask reconstruction, and multi-modal fusion, respectively.}
    \scalebox{0.65}{
     \begin{tabular}{ccccccccccc}
    \toprule
    \multirow{2}{*}{Task}&
    \multicolumn{4}{c}{Module}&
    \multicolumn{3}{c}{ShapeNetRender}&
    \multicolumn{3}{c}{ModelNet}\\
    \cmidrule(lr){2-5} \cmidrule(lr){6-8} \cmidrule(lr){9-11} 
    &CL-F&CL-M&MR&MF &16-bit&32-bit&64-bit&16-bit&32-bit&64-bit \\
    \midrule 
    \multirow{4}{*}{I$\rightarrow$P}
    &\xmark&\xmark&\cmark&\cmark&0.090&0.091&0.090&0.032&0.031&0.031\\
    &\cmark&\xmark&\xmark&\xmark&0.710&0.728&0.749&0.380&0.439&0.471\\
    &\xmark&\cmark&\xmark&\xmark&0.714&0.730&0.748&0.382&0.442&0.473\\
    &\cmark&\cmark&\xmark&\xmark&0.727&0.745&0.760&0.387&0.446&0.477\\
    &\cmark&\cmark&\cmark&\xmark&0.748&0.766&0.784&0.401&0.460&0.492\\
    &\cmark&\cmark&\cmark&\cmark&\bf 0.760& \bf 0.775& \bf 0.791&\bf 0.409&\bf  0.466& \bf 0.501\\
    \midrule
    \midrule
    \multirow{4}{*}{P$\rightarrow$I}
    &\xmark&\xmark&\cmark&\cmark&0.082&0.090&0.081&0.031&0.032&0.032\\
    &\cmark&\xmark&\xmark&\xmark&0.709&0.729&0.745&0.419&0.477&0.511\\
    &\xmark&\cmark&\xmark&\xmark&0.711&0.732&0.746&0.422&0.482&0.513\\
    &\cmark&\cmark&\xmark&\xmark&0.723&0.744&0.758&0.426&0.489&0.521\\
    &\cmark&\cmark&\cmark&\xmark&0.743&0.764&0.781&0.442&0.508&0.537\\
    &\cmark&\cmark&\cmark&\cmark&\bf 0.758&\bf 0.772&\bf 0.790&\bf 0.452&\bf 0.516&\bf 0.545\\
    \bottomrule
    \end{tabular}
    }
    \label{tab:ablation}
    \vspace{-1em}
\end{table}

\section{Conclusion}
In this paper, we propose a self-supervised cross-modal hashing method that addresses the research gap in the field of retrieval between 2D images and 3D point-cloud data. The proposed \ourmethod demonstrates exceptional discriminative capability and robust local perceptibility. The generated hash codes not only capture abundant semantic information but also display a small semantic gap between modalities. The experimental results on three benchmark datasets confirm \ourmethod's superiority. We expect this work can inspire more solid work for research on the cross-modal hashing between 2D data and 3D data in the community and can provide a valuable complement to existing multimedia retrieval systems.


\bibliographystyle{IEEEbib}
\small
\bibliography{reference}

\section{Appendix}
\subsection{Reproduction Guidelines}

\noindent\textbf{Experimental Environment.} The experiments are conducted on a machine with Intel(R) Xeon(R) Gold 6226R CPU @ 2.90GHz, and a single NVIDIA RTX 3090 GPU with 24GB GPU memory. The operating system of the machine is Ubuntu 20.04.5 LTS. For software versions, we use Python 3.8.13, Pytorch 1.12.1, and CUDA 11.7.

\noindent\textbf{Datasets Preparation.} We utilize three publicly available datasets to validate the effectiveness of our proposed method, namely ShapeNetRender, ModelNet, and ShapeNet-55.1) The ShapeNetRender dataset is a subset of the ShapeNet dataset, consisting of 13 categories and containing 43,783 point clouds. To generate the corresponding rendered images, we employ the technique introduced in~\cite{DISN2019NIPS} to randomly select viewpoints. For implementation details, please refer to the GitHub repository\footnote{https://github.com/Xharlie/ShapenetRender\_more\_variation}. 2) The ModelNet dataset comprises 12,311 CAD-generated meshes belonging to 40 categories, such as airplanes, cars, plants, and lamps. The associated point cloud data points are uniformly sampled from the mesh surfaces and subsequently preprocessed by aligning them to the origin and scaling them to fit within a unit sphere. For this dataset, the image-point cloud pairs can be downloaded from the GitHub repository\footnote{https://github.com/zhanheshen/pointcmt} of PointCMT~\cite{PointCMT2022NIPS}. 3) ShapeNet-55 is a publicly available subset of the ShapeNet dataset, consisting of approximately 52.5K 3D objects with 55 annotated categories. Image-point cloud pairs for this dataset can be obtained from the GitHub repository\footnote{https://github.com/salesforce/ULIP} of ULIP~\cite{ULIP2023CVPR}.

\noindent\textbf{Pre-trained Backbones.} In our implementation, we utilize pre-trained backbones to achieve fast convergence with limited training data, following the settings commonly used in most hashing methods~\cite{DGCPN2021AAAI,JDSH2020SIGIR,DJSRH2019ICCV}. For the image branch, we load the parameters of the backbone from the pre-trained DeiT model\footnote{https://github.com/facebookresearch/deit}, which has been trained on ImageNet for image classification. As for the point cloud branch, we employ the pre-trained Point-MAE model\footnote{https://github.com/Pang-Yatian/Point-MAE} as the backbone, which has been pre-trained on the ShapeNet dataset using the masked modeling task.

\begin{table}[t]
    \large
    \centering
    \caption{The number of parameters, FLOPs, and encoding time of each component in \ourmethod.}
   \scalebox{0.6}{
     \begin{tabular}{ccccccc}
    \toprule
    \multirow{2}{*}{Components}&
    \multicolumn{3}{c}{Image}&
    \multicolumn{3}{c}{Point Cloud}\\
    \cmidrule(lr){2-4} \cmidrule(lr){5-7} 
    & Encoder&Fusion&Decoder& Encoder&Fusion&Decoder \\
    \midrule 
    Parameters&86.2M&29.2M&22.4M&22M&29.3M&7.4M\\
    FLOPs&42.2G&1.5G&4.3G&8.9G&1.5G&1.9G\\
    \midrule
    Encoding Time&\multicolumn{3}{c}{10.3ms}&\multicolumn{3}{c}{13.4ms}\\
    \bottomrule
    \end{tabular}}
    \label{tab:complexity}
    \vspace{-1em}
\end{table}

\subsection{More Experimental Results}

\begin{figure*}[thb]
    \centering
  \subfloat[Image codes before training]{\includegraphics[width=0.33\textwidth]{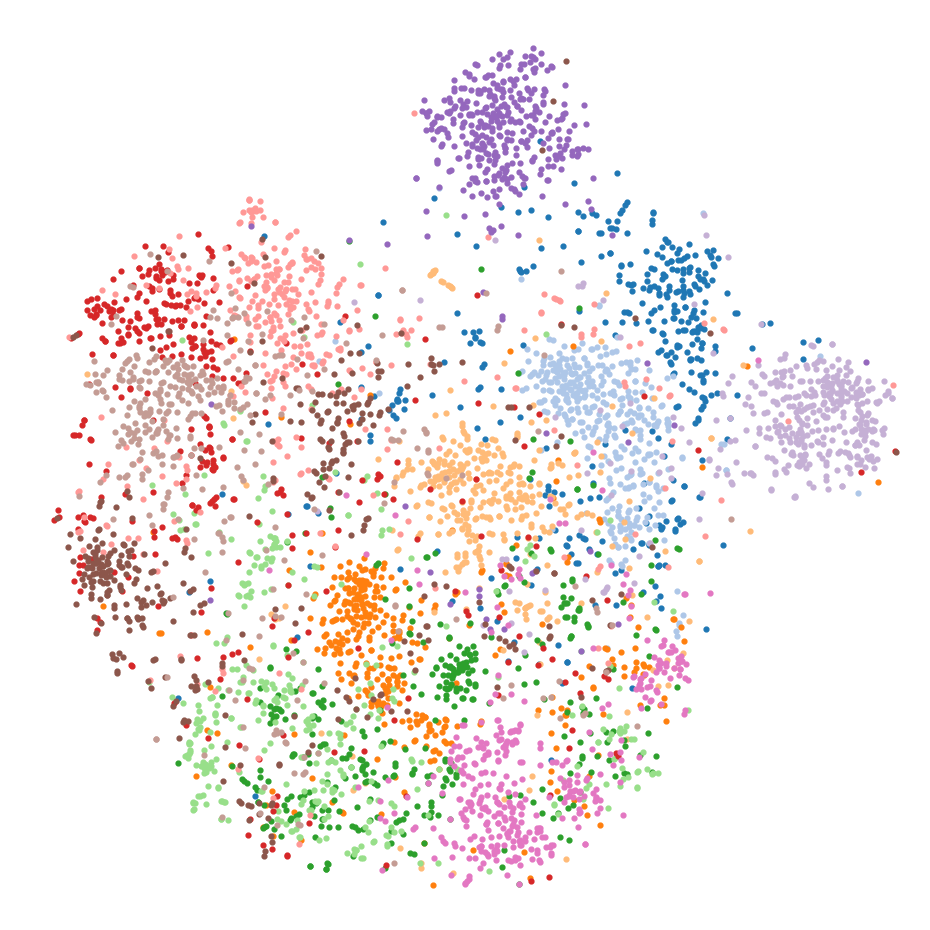}}
    \hfill
    \subfloat[Point cloud codes before training]{\includegraphics[width=0.33\textwidth]{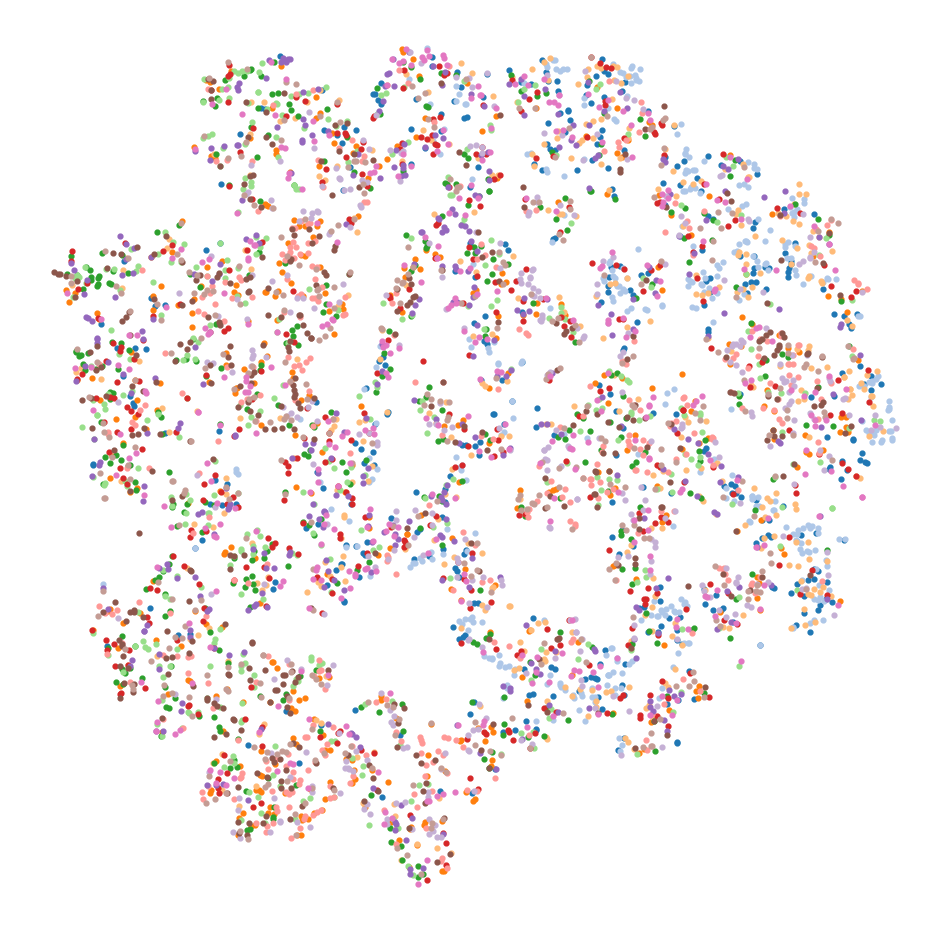}}
    \hfill
    \subfloat[Mixed codes before training]{\includegraphics[width=0.33\textwidth]{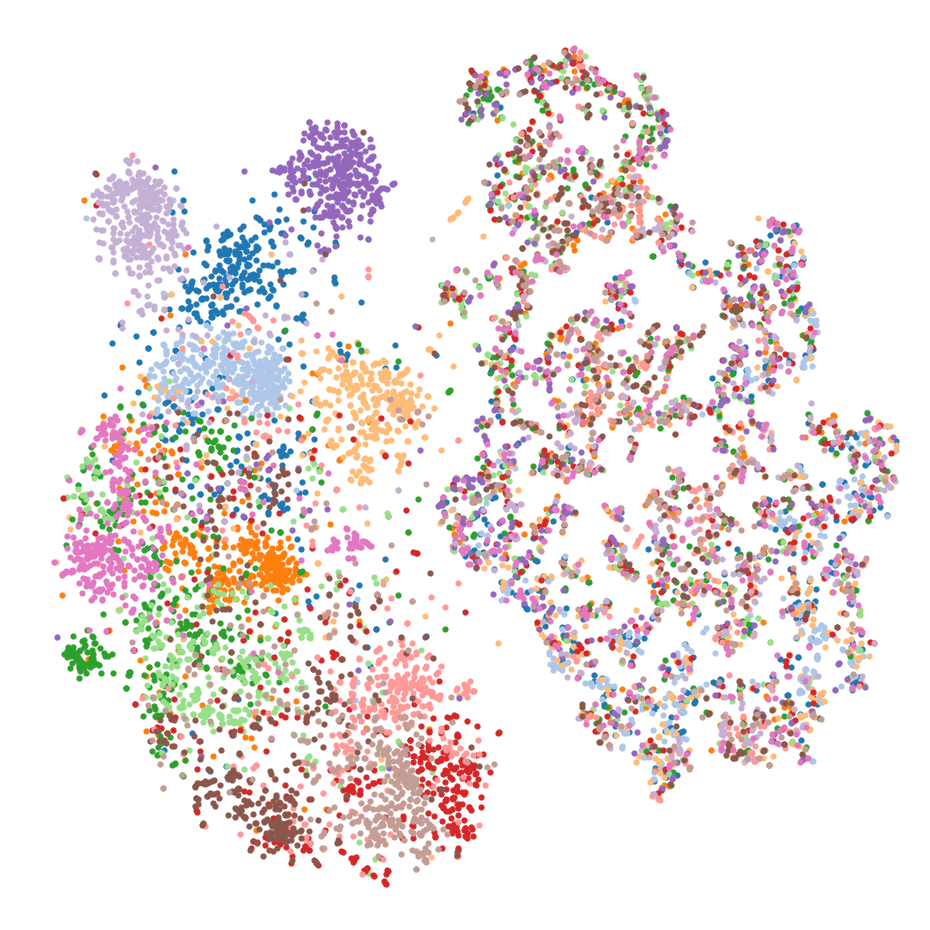}}
    \\
    \subfloat[Image codes after training]{\includegraphics[width=0.33\textwidth]{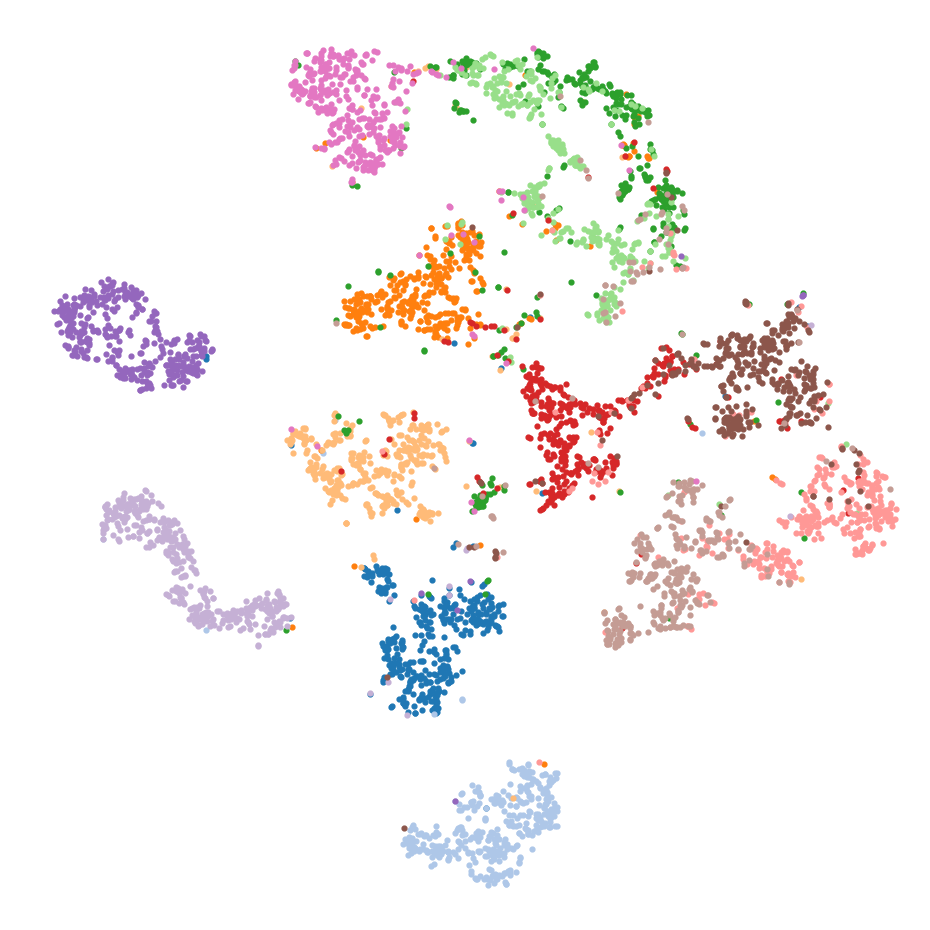}}
    \hfill
    \subfloat[Point cloud codes after training]{\includegraphics[width=0.33\textwidth]{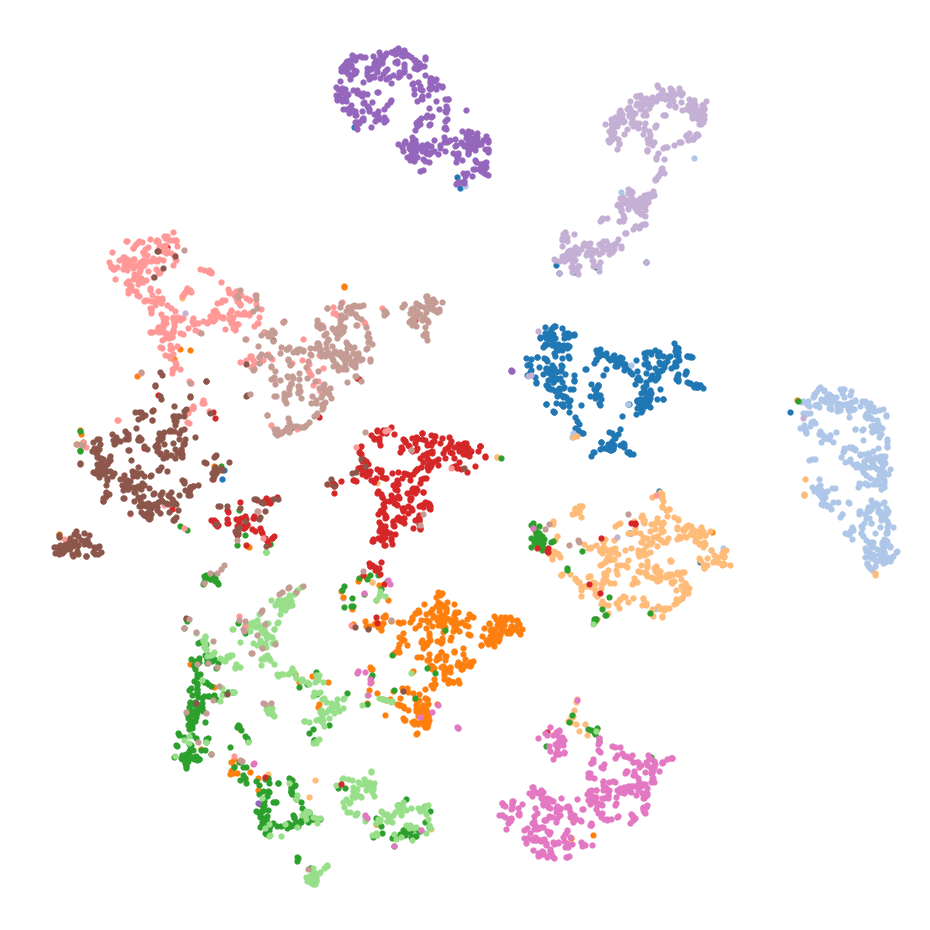}}
    \hfill
    \subfloat[Mixed codes after training]{\includegraphics[width=0.33\textwidth]{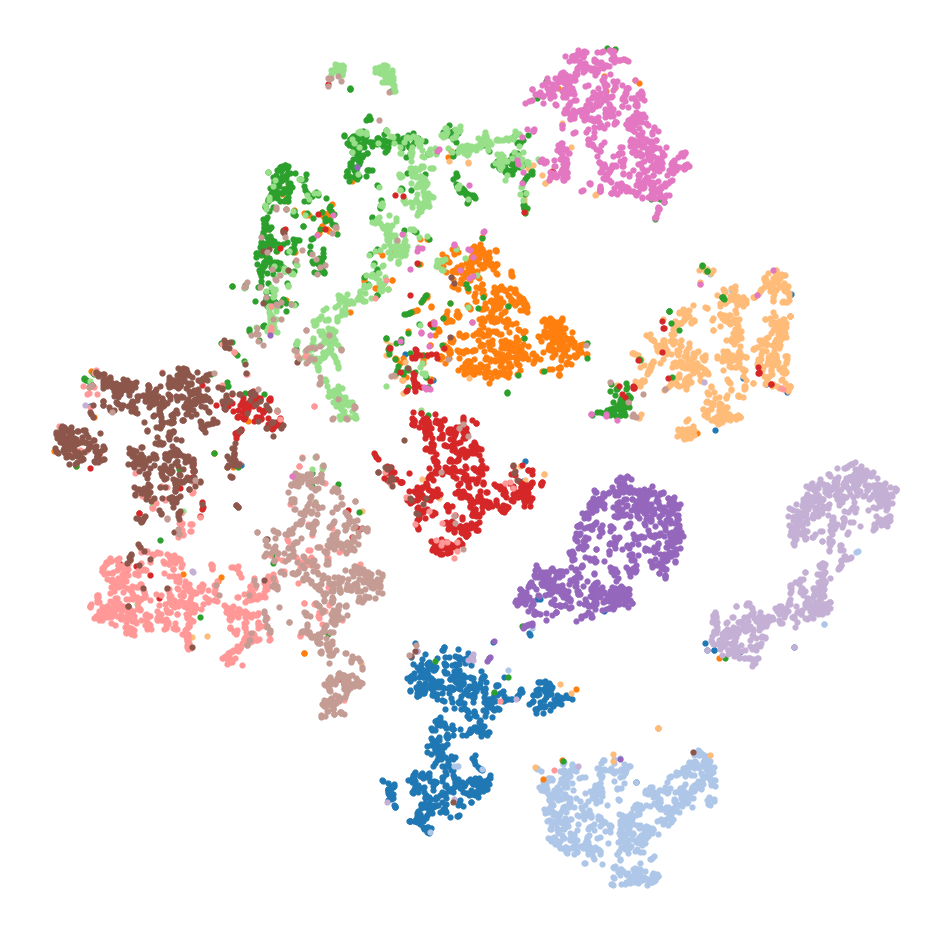}}
    \caption{The t-SNE visualization for 64-bit hash codes before training and after training on ShapeNetRender dataset.}
    \label{fig:t-sne}
    \vspace{-1em}
\end{figure*}

\noindent\textbf{Hash Codes Visualization.} Following~\cite{xinluo2023TCSVT,xiaoluoICME,DUMCH2023ICME}, we visualize the hash codes using the T-SNE algorithm in Figure~\ref{fig:t-sne}. We can observe from (a) and (b) that both the image and point cloud hash codes before training appear messy and indistinguishable. Additionally, the mixed codes demonstrate a significant modality gap, as depicted in (c), with image codes and point cloud codes widely separated and each group gathering independently. However, after training with \ourmethod, it becomes evident that both the image codes and point cloud codes are distinguishable, as shown in (d) and (e). They exhibit distinct compactness for the same category and dispersion for dissimilar ones. Most importantly, as depicted in (f), it is clear that image and point cloud codes with the same semantics can be gathered closely. This visualization demonstrates that our method can generate high-quality hash codes with negligible modality gaps.

\noindent\textbf{Complexity Analysis.} We analysis the complexity of each components in \ourmethod, including the number of parameters, FLOPs, and encoding time. The results are reported in Table~\ref{tab:complexity}. Firstly, we can observe from the table that image encoder takes up most of the model parameters. Decoders have less parameters since they consist of  fewer transformer blocks. As a consequence, image encoder requires large float point operations compared with other components.

Besides, we also evaluate the encoding time (\ie, the time cost from raw image/point cloud to binary code) of \ourmethod in the third row. The measurements are based on the average time taken to encode 100 images/point clouds using a single NVIDIA RTX 3090 GPU. Remarkably, \ourmethod exhibits remarkable efficiency, requiring merely 10.3ms and 13.4ms on average to encode an image and a point cloud, respectively. These results affirm that \ourmethod is highly suitable for integration into multimedia retrieval systems, delivering prompt response within an acceptable latency.

\end{document}